\documentclass[10pt,conference,a4paper]{IEEEtran}
\usepackage{algorithm}
\usepackage[noend]{algpseudocode}
\usepackage{color,array}
\usepackage{xcolor}
\newcommand{\MS}[1]{{\color{blue}  #1}}
\newcommand{\YSR}[1]{{\color{red}  #1}}
\usepackage{times}
\usepackage{epsfig}
\usepackage{graphicx}
\usepackage{amsmath}
\usepackage{amssymb}
\usepackage{footnote}
\definecolor{bg}{rgb}{0.66, 0.66, 0.66}
\definecolor{vtl}{RGB}{189,198,255}
\newcommand{\vtl}[1]{{\colorbox{bg}{\color{vtl}  #1}}}
\definecolor{vrev}{RGB}{1,255,254}
\newcommand{\vrev}[1]{{\colorbox{bg}{\color{vrev}  #1}}}
\definecolor{vstrt}{RGB}{255,238,232}
\newcommand{\vstrt}[1]{{\colorbox{bg}{\color{vstrt}  #1}}}
\definecolor{talk}{RGB}{255,0,246}
\newcommand{\talk}[1]{{\colorbox{bg}{\color{talk}  #1}}}
\definecolor{opndr}{RGB}{107,104,130}
\newcommand{\opndr}[1]{{\colorbox{bg}{\color{opndr} #1}}}
\usepackage{cite}

\ifCLASSINFOpdf
\else
\fi

\begin{document}
%
\title{Gabriella: An Online System for Real-Time Activity Detection in Untrimmed Security Videos}

\author{
\IEEEauthorblockN{Mamshad N Rizve\IEEEauthorrefmark{1}, Ugur Demir\IEEEauthorrefmark{1}, Praveen Tirupattur\IEEEauthorrefmark{1}, Aayush J Rana\IEEEauthorrefmark{1}, \\Kevin Duarte\IEEEauthorrefmark{1}, Ishan Dave\IEEEauthorrefmark{1}, Yogesh S Rawat\IEEEauthorrefmark{2} and Mubarak Shah\IEEEauthorrefmark{2}}
\IEEEauthorblockA{\textit{Center for Research in Computer Vision}\\\textit{University of Central Florida, Orlando, Florida, USA}\\
Email: \IEEEauthorrefmark{1}[nayeemrizve, ugur, praveentirupattur, aayushjr, kevin, ishandave]@knights.ucf.edu, \IEEEauthorrefmark{2}[yogesh, shah]@crcv.ucf.edu}}








\maketitle

\begin{abstract}

Activity detection in security videos is a difficult problem due to multiple factors such as large field of view, presence of multiple activities, varying scales and viewpoints, and its untrimmed nature. The existing research in activity detection is mainly focused on datasets, such as UCF-101, JHMDB, THUMOS, and AVA, which partially address these issues. The requirement of processing the security videos in real-time makes this even more challenging. In this work we propose Gabriella, a real-time online system to perform activity detection on untrimmed security videos. The proposed method consists of three stages: tubelet extraction, activity classification, and online tubelet merging. For tubelet extraction, we propose a localization network which takes a video clip as input and spatio-temporally detects potential foreground regions at multiple scales to generate action tubelets. We propose a novel Patch-Dice loss to handle large variations in actor size. Our online processing of videos at a clip level drastically reduces the computation time in detecting activities. The detected tubelets are assigned activity class scores by the classification network and merged together using our proposed Tubelet-Merge Action-Split (TMAS) algorithm to form the final action detections. The TMAS algorithm efficiently connects the tubelets in an online fashion to generate action detections which are robust against varying length activities. We perform our experiments on the VIRAT and MEVA (Multiview Extended Video with Activities) datasets and demonstrate the effectiveness of the proposed approach in terms of speed ($\sim$100 fps) and performance with state-of-the-art results. The code and models will be made publicly available. 

\end{abstract}


%
\IEEEpeerreviewmaketitle

\section{Introduction}


Deep convolutional neural networks have achieved impressive action classification results in recent years \cite{tran2015learning,carreira2017quo,tran2018closer}. Similar advancements have been made for the tasks of action detection in {\em trimmed videos} \cite{kalogeiton2017action,singh2017online,duarte2018videocapsulenet}, where the spatial extents of the actions are estimated, and {\em temporal} action localization in {\em untrimmed videos} \cite{yeung2015every,piergiovanni2019tempGMM}, where only the temporal extents of the activities are predicted. However, these improvements have not been transferred to action detection in untrimmed videos, where both the spatial and temporal extents of the activities must be found; current approaches have yet to achieve high performance on this difficult task.

\begin{figure}[t!]
\begin{center}
\includegraphics[width=0.9\linewidth]{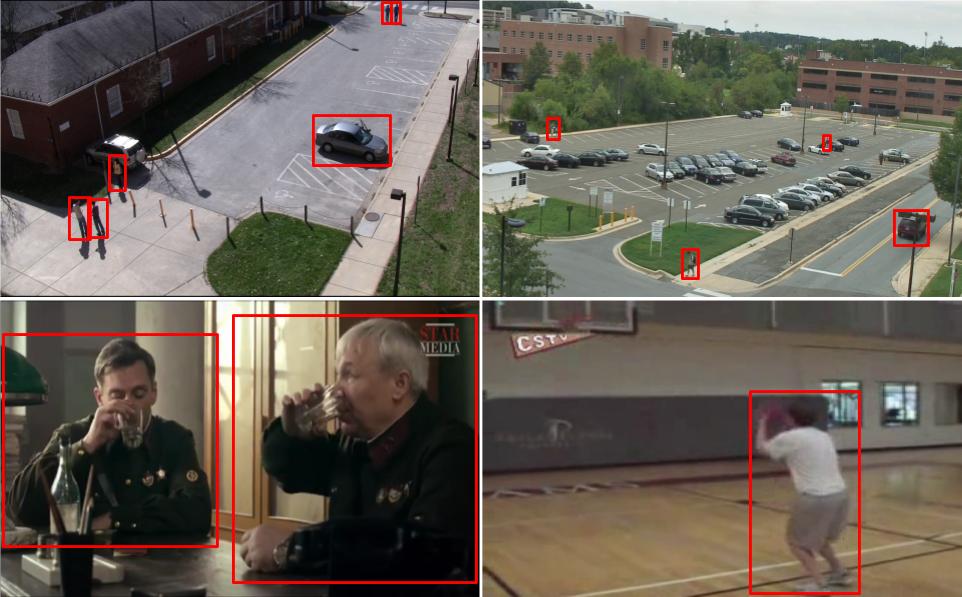}
\end{center}
\caption{\textbf{Top}: Two Sample frames from different scenes of the VIRAT dataset showing variation in perspective, scale and field-of-view. \textbf{Bottom}: Sample frames from the AVA \cite{gu2018ava} (left) and THUMOS'14 \cite{idrees2017thumos} (right) dataset. The VIRAT dataset contains a greater number of concurrent actions as well as a greater variety of action scales (both spatially and temporally). } 
\label{fig:sample_frames}
\end{figure}

Action detection in untrimmed videos has several major challenges: multiple activity types may occur simultaneously, multiple actors may be present, and the temporal extents of the activities are unknown. Videos in trimmed action detection datasets, like AVA \cite{gu2018ava}, contain multiple actors and activities simultaneously, but each video is trimmed into three second clips which have bounding-box annotations only on the center frame. Untrimmed action detection datasets, like THUMOS'14 \cite{idrees2017thumos}, are comprised of untrimmed videos, but each video contains only one or two actors performing the same action. Although difficult, these datasets do not contain all the aforementioned challenges, which results in action detection methods that perform poorly when evaluated on videos containing all these challenges. Therefore, in this work we focus on the surveillence video datasets: VIRAT \cite{oh2011large} and MEVA (Multiview Extended Video with Activities) \cite{meva2019}. Not only do these two action detection datasets have untrimmed videos with multiple activity types and multiple actors, but also they are comprised of multiple viewpoints and contain several actors performing multiple actions {\em concurrently}. These actors have varying scales and actor sizes tend to be extremely small relative to the video frame, which makes the detection of activities in these datasets extremely difficult. Figure \ref{fig:sample_frames} shows sample frames from the VIRAT dataset and compares them with frames from the THUMOS'14 and AVA datasets.

We focus on untrimmed security videos and propose {\bf Gabriella}, an online real-time system for action detection. Our method is composed of three modules: tubelet localization, tubelet classification, and tubelet merging. Our action localization module generates pixel-level foreground-background segmentations which localize actions in short video clips. These pixel-level localizations are turned into short spatio-temporal action tubelets, which are passed to a classification network to obtain multi-label predictions. After classification, the tubelets must be linked together to obtain the final detections with varying length; to this end, our novel online Tubelet-Merge Action-Split (TMAS) algorithm merges these short action tubelets into final action detections.

Conventional action detection methods make use of pretained, frame-based object detectors to localize all potential actors within the video. Frame-based object detection systems, \cite{gleason2019proposal}, have two main issues: 1) processing each frame independently requires large amounts of computation, which reduces the overall speed of the system and leads to temporally inconsistent detections between adjacent frames, and 2) all objects within the frame are detected, even those which are not performing actions. Our action localization module processes multiple frames simultaneously and only produces tubelets which correspond to possible actions within the video. This results in temporally consistent localizations as well as a reduction in the number of proposals, which drastically increases the speed of the overall system. To improve the accuracy of our localization network, we propose a novel Patch-Dice loss. The original global Dice loss \cite{sudre2017generalised} allows networks to account for large imbalances between foreground and background (which is the case for security videos with very small actors). However, it does not take into account the variation in scale of different foreground objects/actions, which leads networks to focus on only the largest actions. The Patch-Dice loss solves this, allowing our network to localize actions of any scale by computing loss on local neighborhoods of each frame. 

Since activities in untrimmed videos can vary in length, it is necessary to handle both short, atomic activities, like {\em `opening a door'} or {\em `exiting a vehicle'}, as well as long, repetitive actions like {\em `walking'} or {\em `riding'}. To this end, our system processes videos in an online fashion. Once the short tubelets have been localized and classified, our Tubelet-Merge Action-Split (TMAS) algorithm merges them into final action tubes of varying length. By classifying short tubelets and merging them into action tubes, our system successfully detects both atomic and repetitive actions. Also, since each tube can have multiple activities co-occurring simultaneously, the TMAS algorithm splits them to successfully isolate individual activities. Due to the online nature of the TMAS algorithm, as well as the efficiency of the localization network, our system generates action detections at over 100 fps, greatly exceeding the speed of contemporary action detection methods.

Our contributions are as follows:
\begin{itemize}
    \item We introduce {\bf Gabriella}, a \textit{real-time online} system that performs action detection in \textit{untrimmed} security videos at \textbf{$\sim$100} frames per second.
    \item We propose an action localization network, trained using a novel \textit{Patch-Dice loss}, to detect \textit{activity agnostic tubelets} of varying scale which significantly reduces the processing time of the system.
    \item The proposed \textit{TMAS} tubelet merging algorithm efficiently connects the tubelets in an \textit{online} fashion and produces detections which are consistent across time as well as robust against varying length activities.
\end{itemize}

We evaluate the proposed approach on the VIRAT and MEVA datasets, and obtain state-of-the-art results in terms of both speed and performance.  


\section{Related Works}
Convolutional Neural Networks (CNN) have been studied for video analysis and applied successfully for the action recognition problem \cite{tran2015learning, carreira2017quo}. Earlier approaches fuse 2D frame features to extract temporal information\cite{karpathyVidcnn}, while recent works mostly apply 3D convolutions to extract spatio-temporal features simultaneously \cite{tran2015learning, carreira2017quo, feichtenhofer2018slowfast}. The works in \cite{feichtenhofer2018slowfast, simonyan2014twostream} use two stream network architectures to further exploit temporal dependencies.

Most action classifiers expect short trimmed videos, however, this is unrealistic for action recognition in real world security videos. Predicting the temporal extents of actions is necessary for reliable recognition. In \cite{piergiovanni2019tempGMM}, a new layer is proposed to temporally localize activities in videos of MultiThumos dataset \cite{yeung2015every}. 
Most works on spatial action detection rely on a region proposal network \cite{frcnn} to detect multiple objects in each frame and combine them temporally to generate action tubelets \cite{yang2017spatio, peng2016multi}. In \cite{hou2017end}, a 3D CNN network efficiently predicts frame-wise background-foreground segmentation map and extrapolates the action tubes. 
However, such approaches becomes computationally inefficient as the number of proposals grows larger, making it unsuitable for real time performance.

Action detection in untrimmed security videos require to address multiple challenges. In \cite{gleason2019proposal}, a frame level object detection and optical flow based model is proposed to solve  action detection on the VIRAT \cite{oh2011large} dataset. They use hierarchical clustering on all detected objects in a video to obtain tube proposals and use optical flow to perform action classification.
In \cite{liu2020argus} authors also perform frame level object detection followed by tracking to generate proposals. These approaches are computationally expensive and not suitable for online processing.
Recent work by \cite{gleason2020real} improves upon~\cite{gleason2019proposal} to make the system real-time by reducing cluster points per video, reduce subsampling rate, and use GPU accelerated optical flow computation. However, this approach produces too many proposals which ends up affecting the system performance.
Our method uses a 3D CNN network for spatio-temporal action segmentation which produces temporally consistent predictions with fewer proposals.
Additionally, our system processes videos in an online fashion without using computationally expensive methods (region proposal network, tracking and optical flow) and achieves better performance in real-time. 


\section{Methodology}

\begin{figure*}[ht!]
\begin{center}
\includegraphics[width=0.92\linewidth]{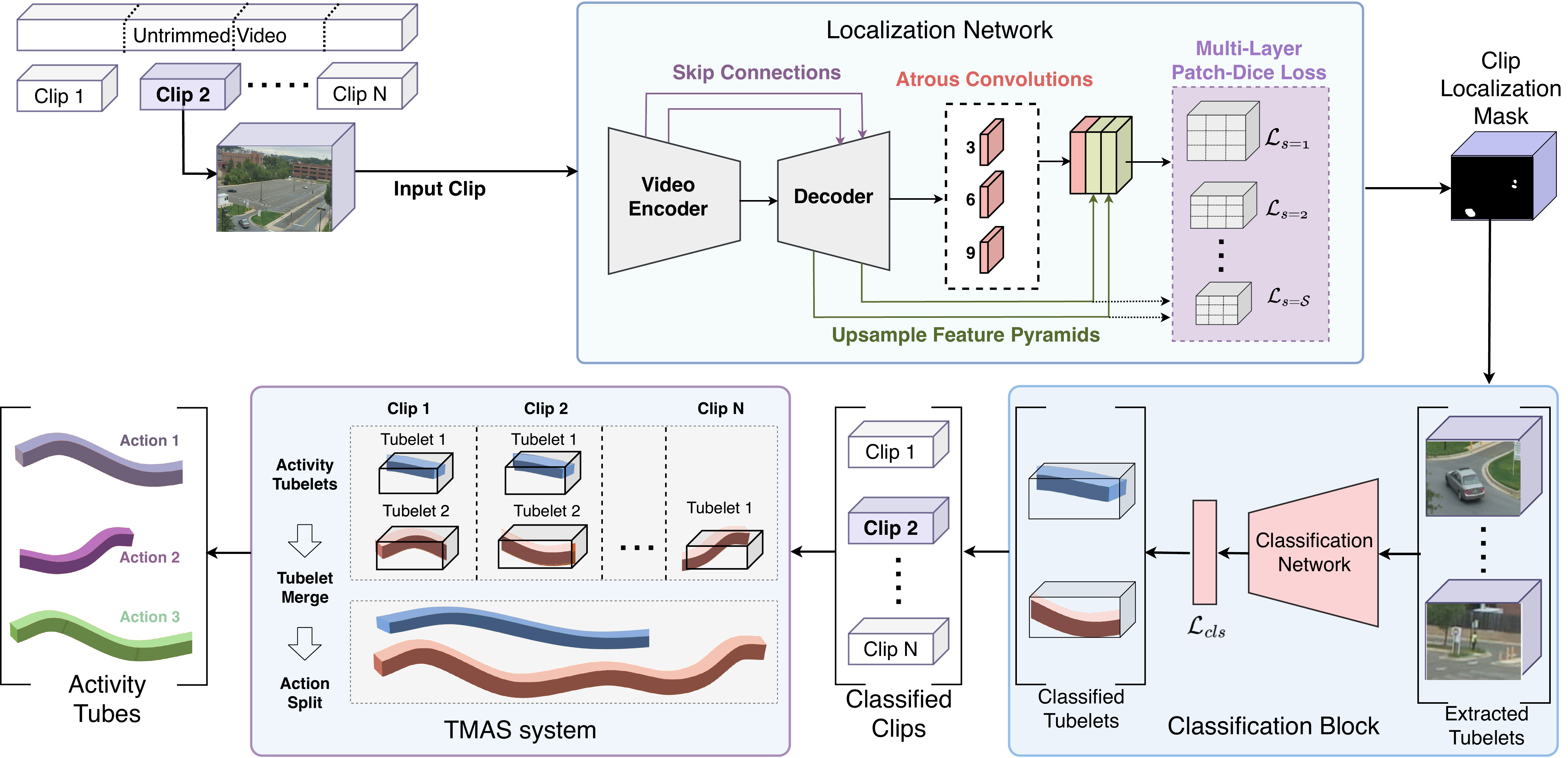}
\end{center}
\caption{Overview of the proposed \textbf{Gabriella} system for action detection in untrimmed videos. An untrimmed video is processed clip by clip and fed into the localization network, producing localization masks. Tubelet extraction produces tubelets for each clip, which are then classified and passed to our TMAS system. The classified tubelets are merged to create action-agnostic tubes, from which individual action-specific detections are obtained. 
}
\label{fig:system}
\end{figure*}

\subsection{Overview}
An overview of our system can be found in Figure \ref{fig:system}. Each untrimmed video is first split into video clips, which are each passed to a localization network to localize potential action tubelets. Then, a network classifies all possible action classes present within each tubelet. Finally, our TMAS system simultaneously filters and combines these short tubelets into longer action tubes. Since our system works on individual video clips in an online fashion, it is able to produce these action detections in real-time. In this section, we will describe the different components of our proposed method.

\subsection{Localization Network} \label{tube_extraction}
The localization network processes a short video clip and localizes all actions within the clip.  The network uses an encoder-decoder structure which extracts class-agnostic action features and generates segmentation masks. We utilize a 3D convolution based encoder, I3D \cite{carreira2017quo}, to extract spatio-temporal features required for activity localization. The decoder must use these features to segment regions from the original input which contain activities. Following recent works in image segmentation \cite{chen2017rethinking, chen2018encoder, kirillov2019panoptic} and video segmentation \cite{duarte2018videocapsulenet, hou2017end}, we use a decoder structure which combines transpose convolutions and upsampling. Stacking many transpose convolution layers is computationally intensive, so we interleave upsampling operations to interpolate features. This results in a shallow decoder network, which prevents over-parameterization and avoids overfitting.

To obtain accurate action localizations, our decoder makes use of skip connections \cite{RFB15a}, feature pyramids \cite{kirillov2019panoptic}, and atrous convolutions \cite{chen2017rethinking}. The decoders' input features have been down-sampled by the encoder, so to obtain fine-grained segmentations we utilize skip connections from higher resolution layers of the encoder. Since many of the actors within security videos vary in scale, decoder makes use of feature pyramids: we stack features from various decoder layers (through upsampling) to obtain feature representations at different scales. Furthermore, we apply atrous convolutions to the final feature representation of the decoder so that the decoder may learn the contextual information necessary for action localization.

\textbf{Patch-Dice Loss:} 
The final output of our localization network is a segmentation mask, $\hat{y}$, where each pixel is assigned a probability of being a part of an action. Given the ground-truth foreground/background mask, $y$, the network is trained end-to-end using a sum of two losses. The first is the binary cross-entropy (BCE) loss,
\begin{equation} \label{eq:bce}
\footnotesize{
    \mathcal{L}_\text{BCE}(y, \hat y) = -\dfrac{1}{N} \sum_{i=1}^{N} \left[y_i log(\hat y_i) + (1 - y_i) log(1 - \hat y_i)\right],
    }
\end{equation}
computed over all $N$ pixels. Since the actors tend to be small in security videos, there is a large imbalance between the number of foreground and background pixels, which causes BCE to miss-localize some actoins. A standard approach to remedy this is to use the Dice loss~\cite{sudre2017generalised}; however, we find that the large variation in scale between different foreground objects (actors) leads the network to focus on the larger actions and ignore smaller actions. 

To this end, we propose a Patch-Dice Loss (PDL), 
\begin{equation} \label{eq:pdl}
\footnotesize{
    \mathcal{L}_\text{PDL}(y, \hat y) = \sum_{k=1}^{K} \left(1-\frac{2\sum_{i=1}^M p_{k,i}*\hat{p}_{k,i}}{\sum_{i=1}^M p_{k,i}^2 + \sum_{i=1}^M \hat{p}_{k,i}^2 + \epsilon} \right)
    }
\end{equation}
where $K$ is the number of patches, $M$ is the number of pixels per patch, and $p_{k,i}$ denotes the probability value assigned to pixel $i$ in patch $k$. This loss splits frames into many local neighborhoods (patches), and computes the dice loss on each patch; this forces the network to segment actions of any size. 

Our final loss is a weighted sum of BCE and PDL, calculated over multiple scales:
\begin{equation} \label{eq:pdl_bce}
\footnotesize{
    \mathcal{L}_\text{loc} = \sum_{s=1}^{S} \lambda_1\mathcal{L}_\text{BCE}\left(y^{(s)}, \hat y ^{(s)}\right) + \lambda_2\mathcal{L}_\text{PDL}\left(y ^{(s)}, \hat y ^{(s)}\right),
    }
\end{equation}
where $S$ denote number of layers and for a layer $s$, $y^{(s)}$ and $\hat y ^{(s)}$ are the ground-truth and predicted segmentations respectively.

\begin{algorithm}[t!]
\footnotesize
\caption{The Tubelet-Merge algorithm which merges tubelets into action-agnostic tubes. The \textproc{CheckEnd} function determines if a candidate tube becomes a final tube or is merged with another candidate.}\label{alg:tubletmerge}
\hspace*{\algorithmicindent} \textbf{Input:} A stream of tubelets, \textbf{S}, from the classifier\\
\hspace*{\algorithmicindent} \textbf{Output:} A set of action-agnostic tubes, $T_{done}$\\
\hspace*{\algorithmicindent} \textbf{Notation: } $\text{Inter}_\text{temp}$ calculates temporal overlap between tubelets. \\
\hspace*{\algorithmicindent} $|\textbf{M}[(\tau_c, *)]|$ returns the cardinality of the set $\{\tau : \textbf{M}[(\tau_c, \tau)] > 0 \}$.
\begin{algorithmic}[1]
\Procedure{Tubelet-Merge}{\textbf{S}}
\State $T_{prev},T_{done} \leftarrow \{\}$ \Comment{Initialize candidate and final tubes}
\State $\textbf{M} \leftarrow \text{initialize hash table}$ 
\While{$\tau_c \text{ in \textbf{S}}$} \Comment{Continue until the stream of tubelets ends}
    \ForAll{$\tau_p \text{ in } T_{prev}$}
        \If{$\text{Inter}_\text{temp}(\tau_p, \tau_c) > 0$}
            \State $\textbf{M}[(\tau_p, \tau_c)] \leftarrow IoU(\tau_p, \tau_c)$
        \Else
            \State \Call{CheckEnd}{$\tau_p$, $T_{prev}$, \textbf{M}}
        \EndIf
    \EndFor
    \State \text{append } $\tau_c$ \text{ to } $T_{prev}$ \Comment{Tubelet becomes a candidate tube}
\EndWhile
\While{$T_{prev} \text{ is not empty}$} \Comment{Deals with remaining candidates}
    \State $\tau_p \leftarrow T_{prev}[0]$
    \State \Call{CheckEnd}{$\tau_p$, $T_{prev}$, \textbf{M}}
\EndWhile
\State \textbf{return} $T_{done}$
\EndProcedure
\end{algorithmic}

\begin{algorithmic}[1]
\Function{CheckEnd}{$\tau_p, T_{prev}, \textbf{M}$}
\If{$|\textbf{M}[(\tau_p, *)]| == 0$}
    \State \Call{Move}{$\tau_p$, $T_{prev}$, $T_{done}$}
    \Comment{Moves $\tau_p$ from $T_{prev}$ to $T_{done}$}
\ElsIf{$|\textbf{M}[(\tau_p, *)]| == 1$}
    \State $\tau_i \leftarrow \max_{\tau_i} \textbf{M}[(\tau_p,\tau_i)]$
    \If{$|\textbf{M}[(*, \tau_i)]| == 1$} 
        \State \Call{Merge}{$\tau_p$, $\tau_i$, $T_{prev}$, \textbf{M}}
    \Else
        \State \Call{Move}{$\tau_p$, $T_{prev}$, $T_{done}$} 
    \EndIf
\Else
    \State $\tau_i \leftarrow \max_{\tau_i} \textbf{M}[(\tau_p, \tau_i)]$
    \State \Call{Merge}{$\tau_p$, $\tau_i$, $T_{prev}$, \textbf{M}}
\EndIf
\EndFunction
\end{algorithmic}

\begin{algorithmic}[1]
\Function{Merge}{$\tau_1, \tau_2, T_{prev}, \textbf{M}$} \Comment{Merges two candidate tubes}
    \State $\tau_1 \leftarrow (f_1^1, f_2^2, \{\textbf{b}^1, \textbf{b}^2\}, \{\textbf{a}^1, \textbf{a}^2\})$ \Comment{$\{\}$ is concatenation}
    \State $\text{remove }  \tau_2 \text{ from } T_{prev}$
    \State $\textbf{M}[\tau_1,\tau_i] \leftarrow \textbf{M}[\tau_2,\tau_i]$ \Comment{$\text{Done for all } \tau_i \text{ where } \textbf{M}[\tau_2,\tau_i] \ge 0$}
\EndFunction
\end{algorithmic}

\end{algorithm}

\begin{algorithm}[t!]
\footnotesize
\caption{The Action-Split algorithm which converts the action-agnostic tubes into action-specific predictions.}\label{alg:actionsplit}
\hspace*{\algorithmicindent} \textbf{Input:} A set of action-agnostic tubes, $T$, and a set of actions, $C$\\
\hspace*{\algorithmicindent} \textbf{Output:} A set of action-specific tubes, $A_G$\\
\hspace*{\algorithmicindent} \textbf{Notation:} The hyperparameters $\kappa, \alpha, \beta, \text{ and } \gamma$ are described in the \\
\hspace*{\algorithmicindent} supplementary materials. $a^i_c[f]$ and $\tau_i[f]$ contain the action prediction 
\hspace*{\algorithmicindent} scores and tube information at frame $f$, respectively.
\begin{algorithmic}[1]
\Procedure{Action-Split}{$T$}
\State $A_G \leftarrow \{\}$ \Comment{Initializes the action-specific tubes}
\ForAll{$\tau_i \text{ in } T$}
    \State $\tau_{smooth} \leftarrow \text{SMOOTH}(\tau_i)$
    \ForAll{$c \text{ in } 1:C$} \Comment{Loop through each action class}
        \State $a_L \leftarrow \text{EXTRACT}(\tau_{smooth}, c)$
        \State $\text{append }  a_L \text{ to } A_G$
    \EndFor
\EndFor
\State $\textbf{return } A_G$  
\EndProcedure
\end{algorithmic}

\begin{algorithmic}[1]
\Function{Smooth}{$\tau_i$} 
    \ForAll{$f \text{ in } f^i_1:f^i_2$}
        \State $a^i_c\left[f\right] \leftarrow \frac{1}{2\kappa + 1} \sum_{k=-\kappa}^{\kappa}{a^i_c\left[f + k\right]}$
    \EndFor
    \State \textbf{return} $\tau_i$
\EndFunction
\end{algorithmic}

\begin{algorithmic}[1]
\Function{Extract}{$\tau_i, c$} \Comment{Extracts tubes of a specific class}
\State $A_L, a_l \leftarrow \{\}$ \Comment{Initialize extracted action tubes and a placeholder} 
\State $count \leftarrow 0$
\ForAll{$f \text{ in } f_1^i:f_2^i$}
    \If{$a_c^i[f] > \alpha$} \Comment{Continue current action tube}
        \State $\text{append }  \tau_i[f] \text{ to } a_l$
        \State $count \leftarrow 0$
    \Else
        \State $count \leftarrow count + 1$
    \EndIf
    
    \If{$count > \beta$} \Comment{Current action tube is finished}
        \State $\text{append }  a_l \text{ to } A_L$
        \State $a_l \leftarrow \{\}, count \leftarrow 0$
    \EndIf
\EndFor
\State $\text{remove tubes shorter than } \gamma \text{ from } A_L$
\State $\textbf{return } A_L$
\EndFunction
\end{algorithmic}

\end{algorithm}

\textbf{Tubelet Extraction:} The segmentation output for each clip is a 
foreground-background confidence mask which isolates potential action tubes. To obtain individual tubelets from this segmentation output, we threshold the output to create a binary mask followed by spatio-temporal connected component extraction. The connected component \cite{Wu2005optimizing, fiorio:inria-00549539} process will generate tubelets for all spatially and temporally linked pixels.

\subsection{Classification Network} Once the tubelets are extracted, our system assigns it action labels. Our classification network is an R(2+1)D network \cite{tran2018closer}, that generates $C+1$ probability outputs, where $C$ is the number of action classes and the additional output is for the background class (used in the case where no action is present in the tubelet). Since multiple actions could occur simultaneously in one tube, we treat this as a multi-label classification problem. Instead of using a softmax activation for the probability outputs, which is common for single-label action classifiers, a sigmoid activation is used which allows multiple actions classified within a single tubelet. To train this classifier, we use a BCE loss (equation \ref{eq:bce}) summed over all $C+1$ probability outputs.

\subsection{TMAS Algorithm}
To merge the tubelets and obtain the final action detections (tubes), 
we propose the Tubelet-Merge Action-Split algorithm (TMAS). Each tubelet, denoted $\tau_i$, is described as follows: $\left(f_1^i, f_2^i, \textbf{b}^i, \textbf{a}^i_c\right)$ where $f_1^i$ is the start time, $f_2^i$ is the end time, $\textbf{b}^i$ are the bounding boxes for each frame of the tubelet, and $\textbf{a}^i_c$ are the frame-level action probability 
$c\in \{ 0, 1, ... C\}$, where $0$ is background. The TMAS algorithm consists of two steps. First, we merge the tubelets into action-agnostic tubes of varying length; then, we split these action-agnostic tubes into a set of action-specific tubes which contain the localizations for the various activities in the video.

\textbf{Tubelet-Merge:} 
The procedure to merge tubelets into action-agnostic tubes is described in Algorithm \ref{alg:tubletmerge}. The input to this is a temporally sequential stream of tubelets coming from the classification network. The set of candidate tubes is initialized with the first tubelet. For each subsequent tubelet, we calculate the spatio-temporal overlap with the existing candidate tubes. This results in four possible outcomes: 1) if there is no overlap, the tubelet becomes a new candidate tube, 2) if there is overlap between a single candidate tube and the tubelet, they are merged and become a new candidate tube, 3) if the tubelet has an overlap with multiple candidates, then the tubelet becomes a new candidate, 4) if multiple tublets have an overlap with a single candidate tube, then the tubelet with the highest overlap is merged with that candidate and the other tubelets become separate candidate tubes. Once all tubelets are checked, the candidate tubes become the output action-agnostic tubes.

\textbf{Action-Split:}
From the action-agnostic tubes we obtain action-specific detections using the Action-Split procedure described in Algorithm \ref{alg:actionsplit}. We start by smoothing out per-frame action confidence scores; which accounts for fragmentation caused by action miss-classifications. Then we build the action-specific tubes by checking for continuous occurrences of each action class; this allows several occurrences of the same activity to occur within a single tube. For instance, a person \textit{walking} might stop and \textit{stand} for several seconds and start walking again; this entire sequence will be contained in a single tube, but the Action-Split procedure will correctly generate two separate instances of \textit{activity\_walking} and one instance of \textit{activity\_standing}. To be robust to classification errors, action tubes with the same action label that are within a limited temporal neighborhood are combined together to form a single continuous action prediction.  

\textbf{Runtime Complexity:}
The worst-case runtime of our TMAS algorithm is $\mathcal{O}\left(n^2\right)$, where $n$ is the total number of candidate tubes at any given time. However, we sequentially process our tubelets and constantly shift the candidate tubes which can not have any possible future match to the set of final tubes. Therefore, the set of candidate tubes at any particular time is reasonably small and our TMAS algorithm contributes negligible overhead to our system's overall computation time.   

\section{Experimental Setup}

\textbf{Datasets:} We evalaute our method on two large-scale action detection datasets with untrimmed security videos: VIRAT and MEVA. The first dataset consists of videos from the VIRAT \cite{oh2011large} dataset with added action detection annotations. It contains 64 videos (2.47 hours) for training and 54 videos (1.93 hours) for validation, with annotations for 40 different activities involving people and vehicles. There is also a held-out test set containing 246 videos (10.11 hours) whose annotations are not made public. 
The MEVA dataset \cite{meva2019} consists of 1056 annotated videos, each 5 minutes long, covering both indoor and outdoor scenes. We use 936 of these videos for training and of the remaining 120 we use 50 for validation and 70 for local evaluation. These videos are annotated with 37 different activities, mainly involving humans and vehicles.
These annotations follow long-tail distribution, i.e, there are few activities which have many annotated instances as they occur very frequently and there are many activities which have very few instances as they are rare.
For the final testing, the system is submitted to an evaluation server where a set of sequestered videos are used for evaluation of the system. More information about the sequestered data and testing protocol for MEVA can be found at ActEv \footnote{https://actev.nist.gov/sdl} website.

\subsection{Training Details}

\textbf{Network Training} 
The videos for both datasets are high resolution, so we rescale the videos to a lower resolution of $800 \times 448$, while maintaining the aspect ratio. The localization network uses a stack of $16$ frames to obtain the binary segmentation masks; the ground-truth for these masks is the bounding box annotations for all actions present within the given frames (regions within the bounding boxes are considered foreground and other regions are considered background). The network is trained using SGD \cite{robbins1951stochastic}, with a learning rate of 1e-3 for about $100000$ iterations. For our BCE+PDL training we have set both the values of $\lambda_1$ and $\lambda_2$ to $1$. Our classification model is trained with a clip length of 16 frames (with a skip rate of 1 to obtain a second long clip) and a spatial-resolution of $112\times112$. For the classifier, we use the ADAM optimizer \cite{kingma2014adam} with a learning rate of 1e-4 for $75000$ iterations on two NVIDIA GeForce Titan X GPUs. 

\textbf{Data augmentation}
To increase the diversity of data, we pre-process the videos which are input to the network during training. For the localization network, we apply frame jitter and cropping to simulate shaking of a camera which can happen due to wind. 
For the classification network, we perform cropping, resizing, and horizontal flipping on the input tubes. Moreover, we use both ground-truth and predicted (outputs of the localization network) tubes for the training of the classifier.

One of the challenging issues with both the VIRAT and MEVA datasets is data imbalance. To balance the data, we first under-sample the classes with largest number of samples. Also, we perform multi-scale cropping and horizontal flipping on classes with the fewest number of samples. Lastly, we perform frame reversal to generate new clips for complementary pairs of classes such as (\textit{Opening}, \textit{Closing}), (\textit{Loading}, \textit{Unloading}), (\textit{Entering}, \textit{Exiting}), and (\textit{Open\_Trunk}, \textit{Close\_Trunk}) to increase the number of samples for these classes.

\textbf{Metrics} 
We evaluate the performance of our system using several metrics: probability of missed detection at fixed rate of false alarm per minute ($\text{P}_\text{miss}\text{@}\text{R}_\text{FA}$), probability of missed detection at fixed time-based false alarm per minute ($\text{P}_\text{miss}\text{@}\text{T}_\text{FA}$), and partial area under the Detection Error Tradeoff curve (AUDC). These measure the quality of action detections for the action detection task.
To calculate these metrics, a one-to-one correspondence is found between the ground-truth actions and the detected actions; ground-truth actions without a corresponding detection are missed detections, while detections without corresponding ground-truth actions are false alarms. For more detailed explanations of the evaluation metrics as well as evaluation code, we refer to TRECVID-2019 \cite{actev2019} and MEVA SDL \cite{meva2019}. For ablations of the classification network, we use standard multi-label classification metrics: precision, recall, and F1-score.

\section{Results}
In this section we present the evaluation of our overall method as well as ablations on its individual components. 



\subsection{System Evaluation}
\textbf{Testing Environment:}
The proposed system is evaluated on the remote test servers provided by NIST. The testing system expects a program that is compatible with the ActEV Command Line Interface (CLI) protocol \footnote[2]{https://actev.nist.gov/sdl}. CLI protocol helps our system to communicate with the testing environment and to be executed remotely. It runs all of the submitted algorithms and reports the scores on the publicly available leaderboard.   

\textbf{VIRAT:} 
We present our system's performance on the VIRAT test set from TRECVID-2019 leaderboard in Table \ref{tab:phase2_comparison} for temporal action localization. We compare our results with the teams who submitted to the leaderboard and as you can see from the results we outperform every other team on $P_{miss\text{@}0.15}T_{FA}$ and $P_{miss\text{@}0.15}R_{FA}$ metrics. Compared to method MUDSML (the best performing comparison) we achieve similar results (within $0.7\%$) with respect to AUDC, but we achieve over a $9.5\%$ improvement in terms of $P_{miss\text{@}0.15}R_{FA}$.



\begin{table}[t!]
\footnotesize
\begin{center}
\begin{tabular}{|l|r|r|r|}
\hline
\textbf{Team} & $\boldsymbol{P_{miss\text{@}0.15}T_{FA}}$ & $\boldsymbol{P_{miss\text{@}0.15}R_{FA}}$ & \textbf{AUDC} \\
\hline
Fraunhofer & 0.7747 & 0.8474 & 0.8270\\
\hline
vireoJD-MM  & 0.5482  & 0.7284 & 0.6012\\
\hline
NTT\_CQUPT  & 0.5112 & 0.8725 & 0.6005\\
\hline
Hitachi  & 0.5099 & 0.8240 & 0.5988\\
\hline
BUPT-MCPRL & 0.4328 & 0.7491 & 0.5240\\
 \hline
MUDSML \cite{liu2020argus} & 0.3915 & 0.7979 & \textbf{0.4840}  \\
\hline
\textbf{Ours} & \textbf{0.3858} & \textbf{0.7022} & 0.4909 \\
\hline
\end{tabular}
\end{center}
\caption{Temporal localization results on VIRAT test set from TRECVID-2019 leaderboard. All the metrics relate to the miss-rate, so lower values indicate better performance.}
\label{tab:phase2_comparison}
\end{table}

\textbf{MEVA:} We present the results of our system on the MEVA sequestered test set in Table \ref{tab:meva_comparison}. Our method achieves state-of-the-art results in both metrics: improving AUDC by over 3.5\% and $P_{miss\text{@}0.04}T_{FA}$ by over 2\%. Notably, our system outperforms others without the need of pre-trained object detectors for localization or optical flow for classification. 




\begin{figure}[t!]
\begin{center}
\includegraphics[width=0.85\linewidth]{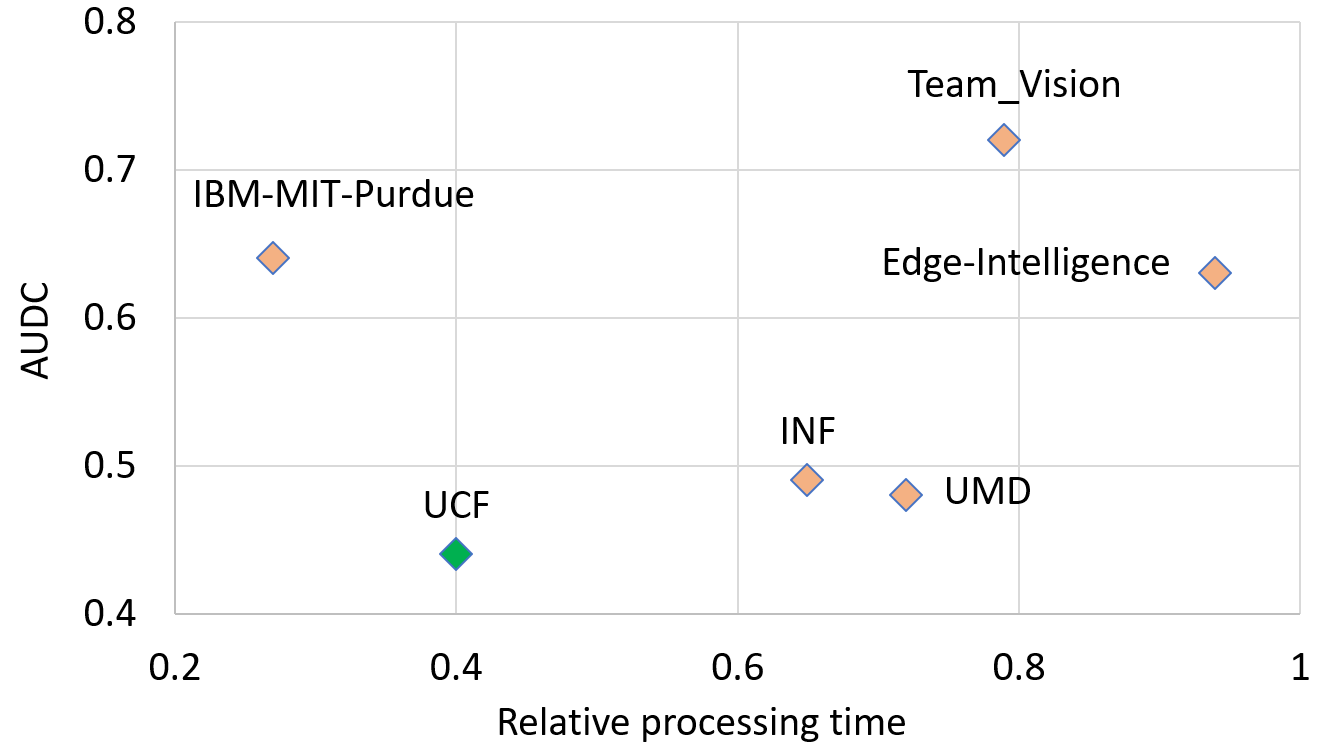}
\end{center}
\caption{Runtime vs AUDC Score of different systems on MEVA test set.} 
\label{fig:runtimePerformanceGraph}
\end{figure}

\begin{table}[t!]
\footnotesize
\begin{center}
\begin{tabular}{|l|r|r|r|}
\hline
\textbf{Team} & \textbf{AUDC} & $\boldsymbol{P_{miss\text{@}0.15}T_{FA}}$ & \textbf{Processing Time}\\
\hline
Team-Vision  & 0.717 & 0.776 & 0.793\\
\hline
IBM-MIT-Purdue  & 0.641 & 0.733 & 0.272\\
\hline
Edge-Intelligence  & 0.628 & 0.754 & 0.939\\
\hline
INF & 0.489 & 0.559 & 0.646\\
 \hline
UMD \cite{gleason2020real}  & 0.475 & 0.544 & 0.725\\
\hline
\textbf{Ours} & \textbf{0.438} & \textbf{0.523} & 0.362\\
\hline
\end{tabular}
\end{center}
\caption{Temporal localization results on MEVA sequestered test set. All the metrics relate to the miss-rate, so lower values indicate better performance. These results are from the publicly available leaderboard \textsuperscript{2}. $P_{miss}R_{FA}$ is not included in this table as it is not made public.}
\label{tab:meva_comparison}
\vspace{-4mm}
\end{table}

\subsection{Run-time analysis}
We compare the speed and performance of our system  with other systems on MEVA test set in Figure \ref{fig:runtimePerformanceGraph}. All systems are tested on 4 NVIDIA RTX 2080 Ti GPUs which is the standard configuration for the evaluation system \cite{actev2019}. Our online action detection method outperforms all most all the other systems by a wide margin. Our method also achieves higher than real-time speed, 45fps, on a \textit{single} GPU. This large difference in speed is mainly due to our localization network: it directly generates tubelets instead of relying on per-frame object detections for proposal generation. This greatly reduces the number action proposals and allows us to process videos online very efficiently.

\begin{figure}[ht!]
\begin{center}
\includegraphics[width=1.0\linewidth]{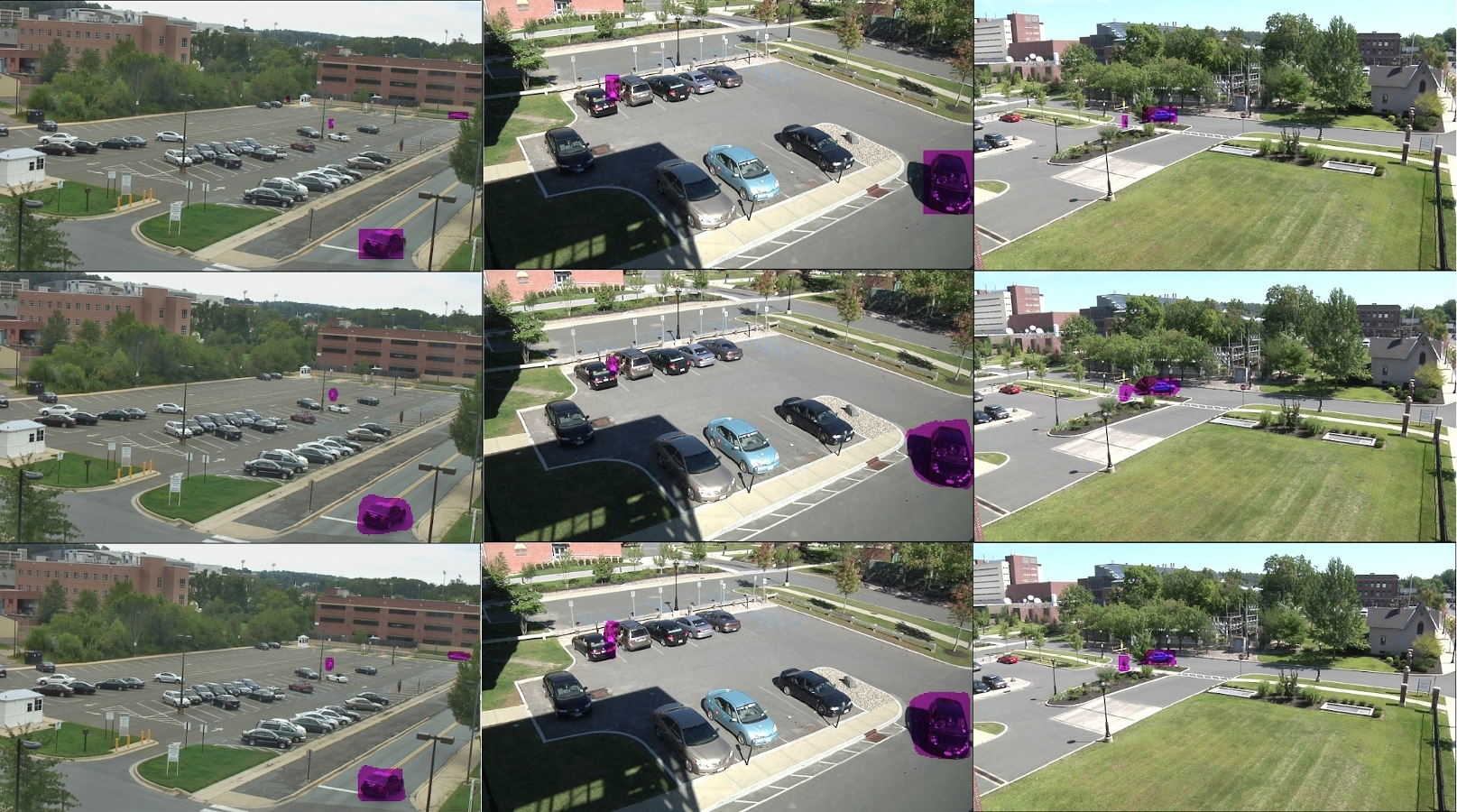}
\end{center}
\caption{Qualitative results from the localization network overlaid on the input frame; the three rows demonstrate action masks obtained from the ground-truth and generated using the BCE, and Patch-Dice loss respectively. The first two columns demonstrate that the network trained with Patch-Dice loss can detect small actions that are missed or partially detected if BCE loss was used. The third column demonstrates that the localization masks generated using the Patch-Dice loss have better action boundaries.}
\label{fig:loc_visualization}
\end{figure}

\begin{table}[t!]
\footnotesize
\begin{center}
\begin{tabular}{|l|c|}
\hline
 \textbf{Model} & \textbf{IoU} \\
\hline
BCE & 62.27\%\\
\hline
BCE + Dice Loss & 62.35\%\\
\hline
BCE + PDL & 63.43\%\\
\hline
\end{tabular}
\end{center}
\caption{Ablation experiments to study the effect of the Patch-Dice Loss on the localization network.  
}
\label{tab:localization_models}
\vspace{-6mm}
\end{table}

\begin{table}[t!]
\footnotesize
\label{}
\begin{center}
\begin{tabular}{|l|r|r|r|r|}
\hline
 \textbf{Architecture} & \textbf{Precision} & \textbf{Recall} & \textbf{F1-Score} \\
\hline
I3D \cite{carreira2017quo} & 0.36 & 0.31 & 0.33  \\
\hline
P3D \cite{qiu2017learning} & 0.43 & 0.41 & 0.41  \\
\hline
3D-ResNet \cite{hara3dcnns} &  0.46 & \textbf{0.43} & 0.44  \\
\hline
R(2+1)D \cite{tran2018closer}  & \textbf{0.50} & \textbf{0.43} & \textbf{0.45}  \\
\hline
\end{tabular}
\end{center}
\caption{Ablation experiments for different classification network architectures. Precision, Recall, and F1-scores are averaged over all classes on VIRAT validation set.}
\label{tab:classification_architectures}
\vspace{-6mm}
\end{table}

\begin{figure*}[ht!]
\begin{center}
\includegraphics[width=1\linewidth]{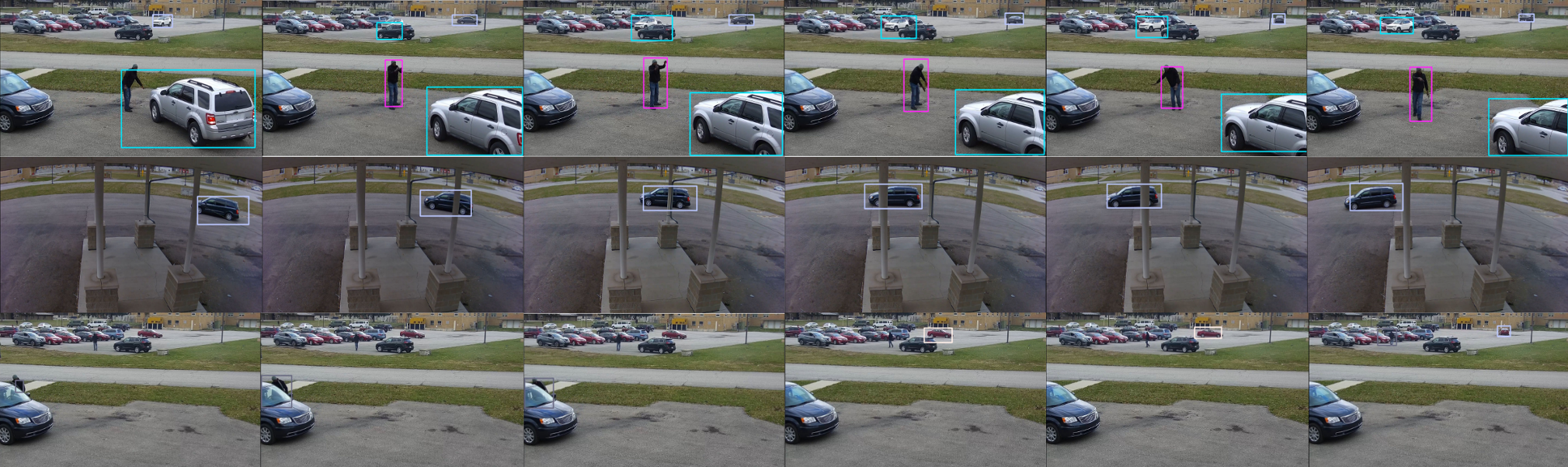}
\end{center}
\caption{Qualitative results of our system on some sample local evaluation videos. Each row is a sample output from our system, showing spatio-temporal localization and classification of actions in specific frames of the input video. Each action type is shown with different colored bounding box. The example activities shown here are \vtl{vehicle turns left}, \vrev{vehicle reverses}, \vstrt{vehicle starts}, \talk{person talks to person}, and \opndr{person opens vehicle door}. These results demonstrate the ability of our system in handling variation of object scales and detecting multiple action classes. 
} 
\label{fig:visualization}
\end{figure*}

\subsection{Ablations}

\textbf{Patch-Dice Loss:}
We run several ablations to evaluate the effectiveness of the Patch-Dice Loss, and present the results in Table \ref{tab:localization_models}. Using PDL during training leads to an improvement in the localization network, mainly due to the increase in number of correct detections. Although the regular dice loss improves localizations when compared to standard BCE, we find that the network does not correctly localize the very small activities. By using PDL during training, the network correctly localizes more of these activities which leads to an overall improvement in the AUDC score. 

\textbf{Classification Network:}
We experiment with multiple classification models to determine the best network architecture for our system. For a fair comparison, all models are initialized with pre-trained weights on the Kinetics \cite{kay2017kinetics} and are trained with the same settings. A comparison of their performance on the VIRAT validation set is shown in Table \ref{tab:classification_architectures}. We use average F1-Score as a metric for comparison and observe that R(2+1)D model \cite{tran2018closer} outperforms the other models. 



\begin{figure}[ht]
\begin{center}
\includegraphics[width=1\linewidth]{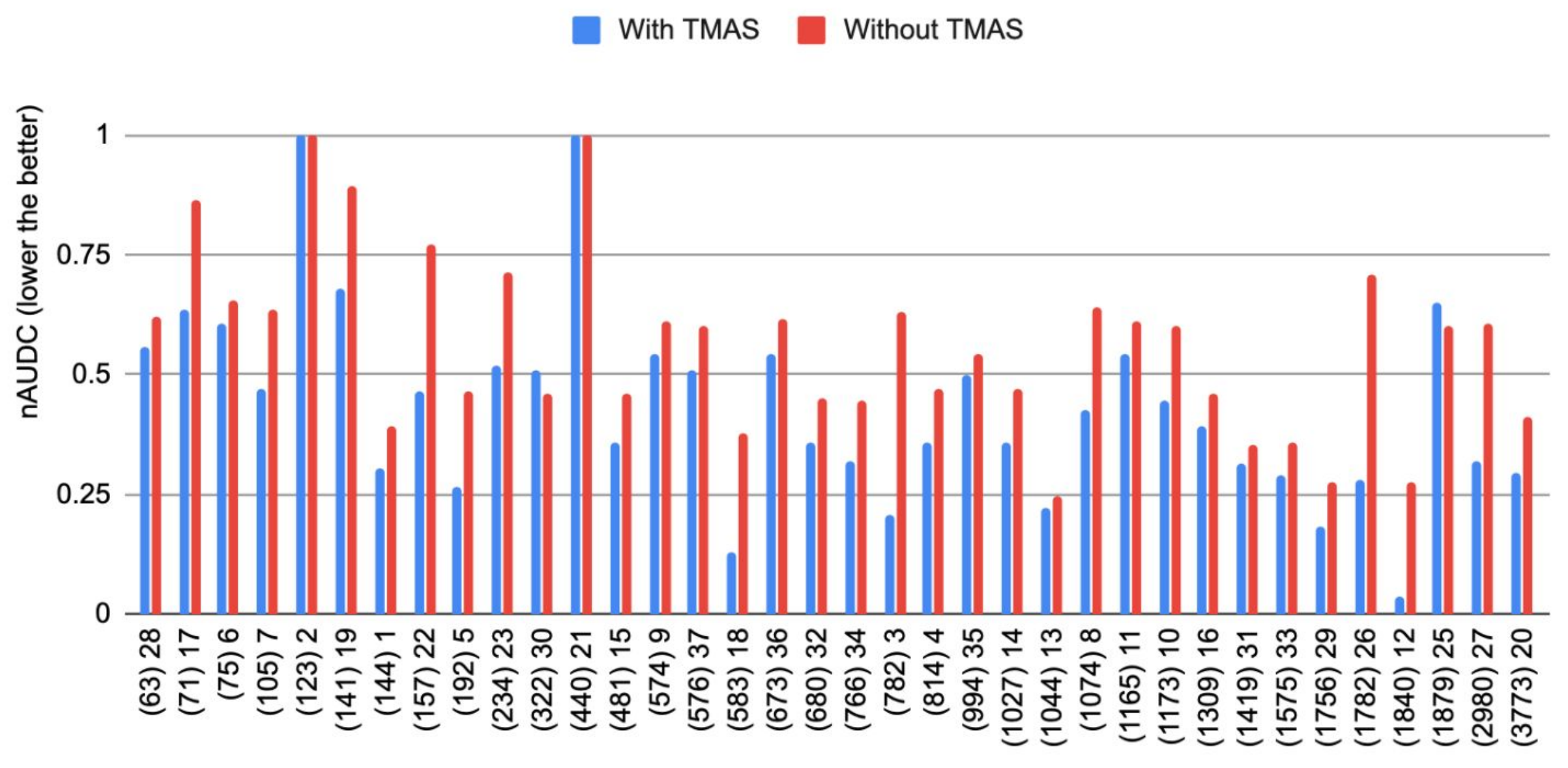}
\end{center}
\caption{Per-class n-AuDC scores from our system, with and without the TMAS. The results show the significance of the Tublet-Merge and Action-Split algorithms in the overall performance of our system. The labels on the x-axis are the class indices and the numbers in the brackets indicate average length of the activity in frames. Please refer to Kitware annotations page 
\textsuperscript{3} for the activity names.}
\label{fig:tmas_ablation}
\vspace{-2mm}
\end{figure}

\textbf{TMAS System}
TMAS is the final step in our system and is crucial for it's success. To show the impact of this step in the overall performance, we compare per-class n-AuDC scores with and without the TMAS algorithm on our local evaluation set of the MEVA dataset. Please refer to Figure \ref{fig:tmas_ablation}. With the post-processing step, we observe that the scores improve for the activity classes which occur for a longer temporal span such as `person reads document'(20) and  `person texts on phone'(24). Please refer to Kitware page \footnote[3]{https://tinyurl.com/rum4ykm} for activity name and the corresponding indices.

\subsection{Qualitative Analysis} 
We present some qualitative results of our system in Figure \ref{fig:visualization}. Our system performs well on different viewpoints, action scales, and action types. We are able to detect activities involving multiple actors, as well as activities involving interactions between a person and a vehicle. Since we do not rely on frame-based object detection, we produce fewer detections and avoid objects which are not involved in an activity - this results in a drastic reduction in computing power used to classify non-actions, like stationary vehicles.

\subsection{Online action detection:}
Online action detection is different from traditional action detection as its goal is to detect an action as it occurs. 
Our proposed system is an online action detection system as it can process a stream of input frames: it perform localization, classification, and temporal segmentation of activities with little or no delay. Other systems, such as \cite{gleason2019proposal} and \cite{liu2020argus} are restricted to offline detection as they rely on object detection for every frame in the video, requiring access to future frames to generate tube proposals. While \cite{gleason2020real} improves computation time, it relies on optical flow computation and produces many proposals, causing trade-off in system performance. This is a major advantage of our system as it can be readily used in real-world security applications.

\section{Conclusion}
In this work, we propose Gabriella, a real-time online system to detect activities in untrimmed security videos. The proposed system consists of three main components which includes tubelet extraction, classification, and online tubelet merging. The proposed approach processes short video clips independently which helps in real-time online processing. The efficient merging of tubelets using TMAS algorithm makes the action detections robust to varying length activities. In contrast to existing approaches, it does not require frame level object detection or optical flow extraction which are computationally expensive and need externally trained models. The proposed method provides state-of-the-art results on the VIRAT and MEVA datasets with a processing speed of over 100 fps. 

\section*{Acknowledgements}
This research is based upon work supported by the Office of the Director of National Intelligence (ODNI), Intelligence Advanced Research Projects Activity (IARPA), via IARPA R\&D Contract No. D17PC00345. The views and conclusions contained herein are those of the authors and should not be interpreted as necessarily representing the official policies or endorsements, either expressed or implied, of the ODNI, IARPA, or the U.S. Government. The U.S. Government is authorized to reproduce and distribute reprints for Governmental purposes notwithstanding any copyright annotation thereon.


{\small
\bibliographystyle{ieee}
\bibliography{egbib}
}



\end{document}